\documentclass[letterpaper,paper,11pt]{AAS}		

\usepackage{bm}
\usepackage{amsmath}
\usepackage[colorlinks=true, pdfstartview=FitV, linkcolor=black, citecolor= black, urlcolor= black]{hyperref}
\usepackage{overcite}
\usepackage{footnpag}			      	
\usepackage{soul}
\usepackage{subcaption}
\usepackage{subcaption}
\usepackage{tcolorbox} 

\PaperNumber{25-426}

\begin{document}

\title{DEVELOPMENT OF A LINEAR GUIDE-RAIL TESTBED FOR PHYSICALLY EMULATING ISAM OPERATIONS}

\author{Robert Muldrow\thanks{Undergraduate Research, MAE, University of Florida, Gainesville, Florida.},  
Channing Ludden\thanks{Graduate Research, MAE, University of Florida, Gainesville, Florida.},
\ and Christopher Petersen\thanks{Principal Investigator Research, MAE, University of Florida, Gainesville, Florida.}
}

\maketitle{}

\begin{abstract}
In-Space Servicing, Assembly, and Manufacturing (ISAM) is a set of emerging operations that provides several benefits to improve the longevity, capacity, mobility, and expandability of existing and future space assets. Serial robotic manipulators are particularly vital in accomplishing ISAM operations, however, the complex perturbation forces and motions associated with movement of a robotic arm on a free-flying satellite presents a complex controls problem requiring additional study. While many dynamical models are developed, experimentally testing and validating these models is challenging given that the models operate in space, where satellites have six-degrees-of-freedom (6-DOF). This paper attempts to resolve those challenges by presenting the design and development of a new hardware-in-the-loop (HIL) experimental testbed utilized to emulate ISAM. This emulation will be accomplished by means of a 6-DOF UR3e robotic arm attached to a satellite bus. This satellite bus is mounted to a 1-DOF guide-rail system, enabling the satellite bus and robotic arm to move freely in one linear direction. This experimental ISAM emulation system will explore and validate models for space motion, serial robot manipulation, and contact mechanics. 
\end{abstract}

\begin{tcolorbox}[colback=gray!10, colframe=black, 
                  arc=1.5mm, boxrule=1.5pt, width=\textwidth]
\normalsize
This is the author’s original manuscript (pre-print) of the paper 
AAS 25-426, presented at the \textit{35th AAS/AIAA Space Flight Mechanics Meeting}, 
Kaua’i, Hawaii, January 19–23, 2025.
\end{tcolorbox}

\section{Introduction}

The emerging capabilities offered by In-Space Servicing, Assembly, and Manufacturing (ISAM) can vastly expand the ranges of operation for in-space assets to improve reusability, mobility, expandability, sustainability, and mission lifespans.\cite{Arney2023} ISAM operations permit servicing of existing satellites, repurposing and recycling of satellites, manufacturing and construction in-orbit, refueling, and upgrades to existing satellites.\cite{Arney2023} Serial robotic manipulators have been proposed as crucial to the continued development of ISAM technologies. Consequently, a 6-DOF robotic arm can be mounted to a free-flying satellite bus, permitting this robotic arm to engage directly with other satellites, space debris, and manufactured parts to accomplish ISAM operations. However, by adding a 6-DOF robotic arm, perturbation forces are introduced to the motion and trajectory of the satellite, even when the robotic arm is not engaged directly with other satellites. This presents three primary challenges; modeling the space motion resulting from conservation of momentum, modeling the serial robotic manipulator, and modeling the contact mechanics when the robotic arm engages with another satellite. This paper endeavors to describe the setup of an experimental testbed capable of emulating these ISAM operations in order to further open questions of multi-body satellite interactions. This linear guide-rail testbed provides a novel experimental setup for simulating ISAM operations. 

Previous use of ISAM has enabled ambitious space missions,\cite{Arney2023} including Northrop Grumman’s MEV,\cite{NorthrupFactSheet} the assembly and continued maintenance of the ISS,\cite{NasaSTS118} and servicing missions to the Hubble Space Telescope.\cite{NasaISAM} With the improvements offered by ISAM, commercial and government space agencies can further missions of human exploration, space logistics, in-space manufacturing and construction, and orbital debris removal, among other advanced missions.\cite{Arney2023}

The challenge of physically emulating and validating resulting space motion and contact mechanics from a robotic arm on a free-flying satellite is a control problem that has been previously studied, however, these experimental testbeds and computational models do not evaluate ISAM operations using a 1-DOF guide-rail system. R. Spicer and J. Black developed a dynamical simulation of a small satellite with a robotic manipulator, however this consideration of robotic manipulators for ISAM only included computational models with no physical, experimental validation.\cite{Spicer2024} Another experimental testbed system by B. R. Fernandez et al. utilizes a tip-tilt air-bearing testbed to approximate proximity-flight orbital mechanics, however this system utilizes two screw actuators to change the orientation of a table surface. Thus, the system uses external effectors to alter the motion of a simulated satellite bus, rather than relying on the actual perturbation forces generated by ISAM. Further, this tip-tilt testbed does not aim to simulate a serial robotic manipulator engaged in ISAM operations, simply to show proximity-flight orbital mechanics.\cite{Fernandez2023} Thus, this linear guide-rail testbed that is presented  provides a novel physical experimental setup to perform simulated ISAM operations with 1-DOF in a linear direction.

This experimental testbed enables direct observation of perturbation forces introduced by the robotic manipulator without the interference of external effectors. By focusing on linear motion, the testbed captures a critical subset of ISAM dynamics, offering a controlled environment to study the interactions between the manipulator, the satellite, and the target objects. This design not only bridges the gap between computational models and physical validation but also lays the groundwork for future studies focused on collecting experimental data to validate theoretical predictions and refine system models. These areas of future studies include three dynamical topics: space motion, serial robot manipulators, and contact mechanics. 

Firstly, this experimental testbed will be capable of examining space motion related to perturbation forces generated by the motion of robotic manipulators. For any given ISAM mission involving movement of manipulators, the motion of manipulators causes an opposing movement by the entire satellite bus, as the center-of-mass (COM) is maintained in accordance with conservation of momentum; the satellite generates a reactionary movement to counter the movement of the robotic manipulator.\cite{Bate1971, Ogilvie2007} The necessity of anticipating such reactionary motions is clear from a review of NASA's Tendon Actuated Lightweight InSpace Manipulators (TALISMAN) mission, in which NASA utilized long-reach robotic manipulators to move payloads.\cite{Talisman2015} If positional error is introduced by these reactionary forces, the robotic arm may be unable to accomplish its task, and thus, reliable space motion dynamical models must account for these reactionary movements.\cite{Sharma2023}

The second area of relevant future research involves updated control of serial robot manipulators. Conventional control of a 6-DOF serial robot manipulator through its revolute joints for the reverse analysis problem is defined with respect to a fixed location,\cite{Crane1998} however, as established above, the system would experience reactionary movements. Thus, the reactionary effects are investigated to update conventional control of robot manipulators.

The final area of relevant future research is contact mechanics, where satellites engaging in Rendezvous, Proximity Operations, and Docking (RPOD) contact each other, resulting in forces that affect the satellites' relative positions and velocities. The initial metrics of interest derive from collision mechanics, a subset of contact mechanics,\cite{Gilardi2002} such as: impulse, coefficient of restitution, and conservation of momentum.\cite{Serway2017} Higher fidelity contact mechanics, such as solving Boussinesq problems with the method of dimensionality reduction (MDR)\cite{Popov2018} are areas of study that can be conducted through this experimental testbed.

With ISAM poised to become a cornerstone of future space operations, understanding the nuanced interplay between robotic manipulators and satellite dynamics is essential.\cite{NasaJscRobo}  This testbed represents the first phase of an ongoing effort to systematically study the mechanics of ISAM operations. At this stage, the focus is on describing the testbed’s construction and design principles, providing a framework for subsequent experimental investigations. This paper is organized as follows. Firstly, the problem setup and goals for the experimental testbed are outlined. Then the designs for specific individual components are discussed along with design choices made. Finally, the manufacturing processes utilized for each component are described, and the results of this construction are reviewed.

\section{Problem Setup}
The objective of this testbed is the emulation of a satellite bus with a mounted 6-DOF robotic arm, where this robotic arm must be capable of freely engaging in a series of simulated ISAM operations. The development of an experimental testbed for ISAM operations depends on the integration of multiple components: the robotic arm, the linear guide-rail system, and the optical breadboard table. Each of these elements is essential to replicating the physical conditions of a satellite equipped with a 6-DOF serial robotic manipulator. However, a fundamental challenge arose during the testbed's construction—these critical components were not inherently compatible in their mounting configurations. Without proper mechanical interfaces, assembling these disparate elements into a cohesive system was not feasible.

To resolve this issue, three custom mechanical interfaces were designed and manufactured: an L-Bracket to mount the robotic arm to the guide-rail, a set of adapter plates to connect the linear guide-rail system to the optical breadboard table, and an 80/20 extrusion table. These interfaces were meticulously engineered to ensure structural integrity, maintain the rigid-body assumption, and align with the precise tolerances required for experimental testing. The rigid-body assumption, namely that all components exhibit negligible deformation in response to operation loads, is necessary to maintain the accuracy and repeatability of experimental results. Without a rigid testbed, small deformations at each interface would accumulate throughout the system, resulting in significant positional and orientation errors for the robotic end-effector, ultimately compromising the accuracy and reliability of the experimental results. Furthermore, a variety of theoretical models used with ISAM make the rigid-body assumption,\cite{WangSpaceRobo2021} and thus, the experimental testbed must align with that assumption to test such models.

In selecting the 6-DOF serial robotic manipulator, the primary consideration was its ability to approximate the motion and mobility of a robotic arm in ISAM operations. While this system provides the necessary degrees of freedom for emulating manipulator dynamics, the testbed design does not account for the size and power constraints typically imposed by space applications. This decision was deliberate, allowing the focus to remain on the mechanical integration and motion modeling challenges unique to ISAM, rather than the limitations of specific hardware.

\section{Robotic Manipulator Selection}

The selection of a robotic manipulator model was a pivotal decision, as this would influence the capabilities of possible emulated ISAM operations, the design and strength requirements of critical components, and the overall integration of the robotic arm to the experimental testbed. 

First, necessary performance metrics were selected, with key metrics being the arm's lift capacity, number of joints, cost, range of speeds, range of motion, total weight, precision and accuracy, and the communication interface. A decision matrix was performed to determine the weight given to each of these metrics in order to quantitatively evaluate a set of commercially available robotic arm models to determine which model to utilize for the testbed.

The robotic arms considered for use with the experimental testbed were as follows: UR3e,\cite{UR3e} UR5e,\cite{UR5e} UR10e,\cite{UR10e} Igus Rebel Cobot, UFACTORY xArm,\cite{xArm} UFACTORY 850,\cite{850Arm} Jaka Zu12,\cite{Zu12} OWI-537,\cite{OWI537} Quanser QArm,\cite{QArm} and KUKA LBR iiwa 7 R800.\cite{LBRiiwa}

The OWI-537 and the Quanser QArm were eliminated, as these failed to meet the 6-DOF specification as stated in the problem statement.\cite{OWI537, QArm} The LBR iiwa 7 R800 exceeded the 6-DOF specification, at 7-DOF,\cite{LBRiiwa} however it was still considered using the decision matrix. The UFACTORY 850 and xArm, as well as the Jaka Zu12,  were eliminated from consideration, as these robotic arms were manufactured in China.\cite{UFactoryOrigin} Thus, using these robotic arms would violate the University's Research Integrity, Security, and Compliance Office import/export criteria.\cite{UFRISC}. This reduced consideration to the UR3e, UR5e, UR10e, Igus Rebel Cobot, and LBR iiwa 7 R800. Based upon the decision matrix and the fact that in-house experience was available for the Universal Robotics line, the UR3e was selected.

\begin{figure}[htb]
	\centering\includegraphics[width=3.5in]{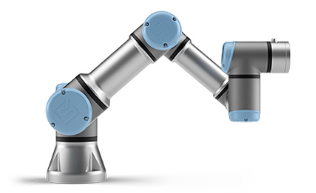}
	\caption{UR3e 6-DOF Robotic Manipulator. \cite{UR3e}}
	\label{fig:UR3eHorizontal}
\end{figure}

The UR3e weighed approximately 24.7 lbs, with a physical footprint diameter of 5.04 inches. As required by the problem statement, the UR3e featured six rotating joints, namely the Base, Should, Elbow, and Wrists 1, 2, and 3. The Wrist 3 joint has infinite working range, while the five other joints only had $\pm 360\deg$. The wrist joints are capable of rotating at $\pm 360 \deg/s$, while the remaining joints have a maximum speed of $\pm 180 \deg/s$. The UR3e is capable of moving a 6.6 lbs payload a range of 19.7 inches. A horizontal image of the UR3e is depicted in Figure \ref{fig:UR3eHorizontal}.

\section{Design Phase 1: L-Bracket}
The first design challenge in constructing the testbed was developing a mounting scheme for horizontally attaching the 6-DOF robotic arm to the guide-rail system. This required a robust mechanical interface capable of providing a rigid connection between the robotic arm and the guide rail while ensuring that the robot manipulator maintained the rigid-body assumption critical for using conventional serial robotic manipulator models. This rigidity was the Key Performance Parameter (KPP) used to assess potential designs. The L-Bracket design selection was driven by minimizing unnecessary complexity. Accordingly, several designs were evaluated that would provide the required rigidity.

\subsection{L-Bracket Designs}
The first design presents the simplest and most cost-effective solution, consisting only of an extruded L-bracket. As noted in Figure 1, multiple sets of holes permit fasteners to connect and secure the robotic manipulator to the guide-rail. 

Design one presented the simplest and most cost-effective solution for the intended objective, consisting of an extruded L-bracket. The rigidity assumption was achieved solely through the thickness and structural integrity of the L-bracket, without any supporting structures. Based on the weight of the UR3e, 24.7 lbs, and the projected maximum payload, 6.6 lbs, the L-bracket thickness must support the exerted momentum caused by these weights without noticeable deflection. Manufacturing for this design would consist primarily of drilling operations required to fasten the L-bracket. Were this design selected, only the thickness of the L-bracket would require consideration.

Design two, three, and four utilized the same general foundation as design one, with the addition of supports to increase rigidity. These designs would require consideration for the position of supports, as well as the thickness of the supports and the L-bracket, the overall size of the supports, and the number of total supports. These supports would be achieved in three separate ways, hence the three additional designs. The second design would achieve these supports through welding. The third design would attach these supports with fasteners. The fourth design would create the supports by removing material from the L-bracket through machining, which would require more specialized stock with excess material in the required locations.

\subsection{Available Manufacturing Tools}

A review of available fabrication and machining resources was conducted to determine which designs were feasible given a limited range of tools and machines. The manufacturing process could be carried out either through custom machining services or using facilities available at the on-campus Design and Manufacturing Laboratory (DML) at the University of Florida. DML offered a variety of equipment, including: Bridgeport vertical mills, manual lathes, Computer Numerical Controlled (CNC) four- and five-axis mills, bandsaws, grinder wheels, a sandblaster, and an abrasive waterjet. Additionally, a variety tools were available, including  High Speed Steel (HSS) and Tungsten Carbide endmills, HSS drill bits, HSS taps, HSS counterbores and countersinks, reamers, dial calipers, parallels, and cylindrical edge finders. Although the shop was equipped with welding tools, these were not publicly accessible at the time of this paper.\cite{DML} Based on general quotes for custom machining services available, external fabrication was deemed cost-inefficient, thus, only DML would be utilized in fabrication of critical components.

\subsection{Design Selection}

Beginning with the fourth design, this was deemed impractical given the highly customized nature of the required stock, which would drastically increase the required cost and limit the range of possible commercial options. Additionally, by creating the supports by removing material with machining, the grain lines of the supports would not align with the natural stress lines, thus decreasing the overall strength and rigidity of the design when compared to a forged part.\cite{CastingForgingDML} Thus, even if the required stock could be obtained, the final results would be sub-optimal.

At the time of this paper, the second design is unfeasible given the limitations of the DML. However, should this public accessibility change, in-house experience with Metal Inert Gas (MIG) welding would enable the production of this design. Additionally, if designs one or three are selected, modification into design two would be relatively simple and cost-effective. Design two presents the strongest and most rigid design, assuming equivalent support specifications and sufficiently welded joints. However, for the time being, this design is only achievable through costly external welding services.

Invariably, the design one was selected, as this presented the simplest possible approach at the present stage. The design one required determination of the acceptable L-bracket thickness and material, while design three would determination of the acceptable thicknesses and materials of L-bracket and support, the size of screws sufficient to support the expected loadings, the size of the supports, the placement of the supports, as well as the total number of supports and of screws. Thus, since design three would substantially complicate manufacturing while providing slightly improved rigidity, the first design was selected for its balance of simplicity, cost-effectiveness, and ease of fabrication, ensuring that it met the required performance specifications within the project's constraints.

\begin{figure}[htb]
	\centering\includegraphics[width=3.5in]{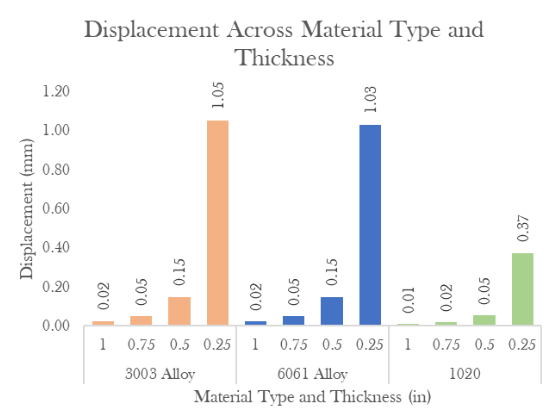}
	\caption{FEA Conducted for 3031, 6061, and AISI 1020.}
	\label{fig:FEA_Study}
\end{figure}

A parametric trade study was conducted to evaluate potential materials, including aluminum alloys (3031, 5032, and 6061) and steel (AISI 10XX series). The 6-DOF robotic arm, modeled using specifications from its technical manual, was represented as a concentrated mass at its center of mass and rigidly connected to the four attachment points outlined in the arm’s user manual. This setup was used to analyze potential deflections under various loading scenarios using SolidWorks simulations and engineering judgment. A series of finite element analysis (FEA) tests were performed across various thicknesses to determine the maximum deflection of the L-Bracket, with results noted in Figure \ref{fig:FEA_Study}.

The analysis suggested that a steel bracket with a thickness of ½ inch would provide the required rigidity without excessive weight or cost. Initial manufacturing investigations explored the feasibility of bending aluminum or steel sheets using a sheet metal brake. However, strain concerns at the bend radius prompted further exploration of welding and subtractive CNC machining options.
Ultimately, due to time constraints and limited access to comprehensive analysis resources, we opted for a pre-bent steel bracket sourced externally. The ½-inch thickness and steel composition were selected based on the FEA results, given that 1/2 inch of a steel material provided a substantial reduction in expected deformation. Further increases in thickness above this point, according to the FEA results, simply provided diminishing returns on the expected decreases in deformation.

This L-bracket design provided a reliable foundation for mounting the robotic arm and satisfied the rigid-body assumption critical for experimental testing. The chosen approach balanced performance, schedule, cost, and producibility, ensuring that the testbed met its design objectives.

\section{Design Phase 2: Adapters}
The design process for the adapter followed a similar methodology to that of the L-bracket, though it presented unique challenges due to its requirement as a "1-plane" device. The primary objective was to create an adapter that would interface with the rail and table components, which had specific geometric constraints, notably the integer spacing of 1 inch for the rail holes, a requirement driven by the use of an optical breadboard. Given these constraints, the adapter needed to accommodate two distinct screw sizes: a ¼-20 screw and an S2-diameter screw with a 5.5 mm diameter, as specified in the rail's technical documentation.

To meet these specifications, the simplest viable design employed bar stock material, which offered an efficient solution for accepting the required screw sizes. The bar stock was processed to incorporate holes that aligned with the different screw hole spacings—1 inch for one type of screw and 30 mm (designated as "C") for the other. This configuration allowed the adapter to accommodate the different spacing requirements while minimizing complexity and manufacturing time.

The adapter design was based on material selection criteria similar to those in Design Phase 1, utilizing aluminum 6061 for its favorable balance of strength, machinability, and cost. Since this component would serve only as a connector, the KPP was that the components were level. If the adapters were not level, fastening them to the rail would actively create deflection. Aluminum 6061 was the ideal material for this goal, as it would be easily machinable. The thickness of the bar stock was determined based on the load-bearing requirements and the available fabrication processes, which ensured that the adapter would perform reliably under the expected mechanical stresses.

This section builds on the previous design considerations, particularly the table section, which required a fixed and rigid interface for the optical breadboard and rail components. As a result, the adapter’s design was constrained by different spacing of the rails vs. the table. The rail had hole-to-hole spacing of 150mm, different than the 1" grid on the table. This meant an asynchronous pattern of relative hole alignments needed to be measured to index the fasteners correctly on the adapter. The use of bar stock allowed for a simple yet effective solution to meet these constraints, with minimal material waste and machining complexity, ensuring that the design could be efficiently fabricated using the available on-campus resources.

\section{Design Phase 3: 80/20 Mounting Table}
The third design phase focused on the structural table, which served as the foundation of the overall assembly. The table design leveraged a commercially available optical breadboard. This breadboard was previously utilized for fluid dynamics and had been repurposed for this experimental testbed. This breadboard, characterized by its precision-machined 1-inch hole spacing, provided a robust and versatile surface for mounting the rail and robotic arm components. The inherent rigidity and flatness of the optical breadboard made it an ideal candidate for this application, minimizing deflection under loading and ensuring reliable alignment across the assembly. 

To secure the optical breadboard within the testbed, 80/20 Aluminum extrusion framing was selected. The 80/20’s modularity and compatibility with the breadboard allowed for straightforward assembly while still providing flexibility for future modifications. The breadboard naturally interfaced with the 80/20 system, eliminating the need for custom adapters or additional hardware. This integration ensured the table could meet the alignment and rigidity requirements of the project.

\section{Manufacturing Phase 1: L-Bracket}
The fabrication of the L-Bracket was conducted using DML facilities with use of the bandsaws and the Bridgeport vertical mills. The extruded ASTM A36 L-Bracket stock was purchased through MetalsDepot, as this offered the most cost effective commercial option. Stock was ordered sufficient to produce two L-Brackets, as this would provide possible replacement stock in the event of manufacturing errors without the necessity of waiting for another delivery. 

\begin{figure}[htb]
	\centering\includegraphics[width=3.5in]{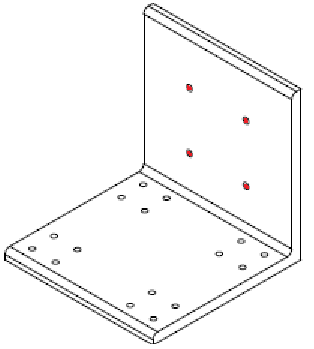}
	\caption{Marked locations on L-Bracket for manufacturing.}
	\label{fig:BracketMarked}
\end{figure}

The bandsaw was utilized to cut the stock into two equal sections of length 8 inches. Then, all edges were broken using a standard file. The L-Bracket stock was then loaded onto the Bridgeport vertical mill, which was utilized to locate and drill all twenty holes required for the design. The workspace constraints required vice grips capable of clamping approximately 8 inches of material. Using this standard setup, sixteen of the twenty holes could be drilled, however, the unique geometry of the L-Bracket prevented the four red holes depicted in Figure \ref{fig:BracketMarked} from being drilled in the configuration shown. This required the L-Bracket be flipped such that the bottom of the part be facing upwards. This left two options, to either find vise jaws capable of clamping onto 8 inches of material or to create soft jaws to hold the L-Bracket in the flipped position. The second option was chosen as the DML did not have any vise jaws of that size.

This process demonstrated the importance of resourcefulness and flexibility in overcoming fabrication challenges within the constraints of available equipment. The choice of materials, the use of soft jaws, and the handling of the part's geometry were all integral to successfully manufacturing the L-Bracket to the required specifications to ensure the L-Bracket maintained the rigid-body assumption and performance requirements.

\section{Manufacturing Phase 2: Adapters}
After suitable designs for the adapters were finalized, Aluminum 6061 bar stock was purchased from McMaster-Carr. The adapter design required several key features, notably precisely drilled holes to accept two screw sizes - $1/4$-20 and M6 x 1.0 - with the former used only as a through hole, while the later fastened directly into the adapter. 

\begin{figure}[htb]
	\centering\includegraphics[width=3.5in]{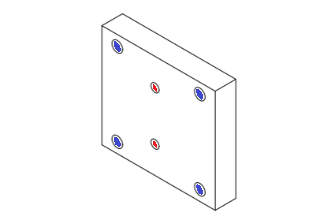}
	\caption{Marked locations on Adapter for manufacturing.}
	\label{fig:AdapterMarked}
\end{figure}

The process of manufacturing the adapters was notably less complex compared to the production of the L-bracket, as the adapters were approximately 2.75" x 2.5" x 0.5". Thus, production of the adapters was well suited for the capabilities of the Bridgeport vertical mill. A cylindrical edge finder was utilized to locate the datums required for the hole locations. The four holes located at the corners of the part, marked in blue on Figure \ref{fig:AdapterMarked}, were drilled with a 0.257" diameter drill as through holes for 5.5mm S2 fasteners. The two holes located in the center of each adapter were drilled with a \#8 drill before being tapped with a M6 x 1.0 tap, with these holes marked in red on Figure \ref{fig:AdapterMarked}. Both diameters were determined using the DML Tap and Drill Chart.\cite{DML}

After production of all eight required adapters was complete, the height of each part was measured using dial calipers. The part with the lowest height was then placed in the mill, and a facing pass was taken such that material was uniformly removed from the top surface of the part. Using the same set of parallels, with the mill's depth in the z-axis remaining constant across all following passes, the other seven adapters were placed in the vice and faced. This ensured that all of the adapters were the same height, within one thousandth of an inch (0.001"). Following this, all eight adapters were sandblasted to produce a professional, matte finish.



\section{Results}
The results from the design and manufacturing process are reflected in a series of evaluations, focusing on the material choices, geometry, and trade-offs that informed the final designs for both the L-bracket and adapter. A set of design charts was developed to visualize key metrics that influenced the decision-making process, including material strength, deflection, weight, and factor of safety. These charts illustrated various configurations of the designs, with each curve representing the relationship between the variables, such as material selection and component thickness. Red circle markers were placed on the curves to indicate the final design choices, which were selected to optimize the balance between structural integrity, manufacturability, and weight. For example, the L-bracket design chart demonstrated the effects of different material thicknesses on deflection under load, highlighting the optimal thickness at which the deflection remained within acceptable limits, ensuring the rigid-body assumption was not compromised.

In parallel, another series of charts was created to evaluate the material, thickness, geometry, and factor of safety for both the L-bracket and adapter. The material selection process for the L-bracket involved an assessment of different alloys, including Al 6061, Al 3031, Al 5032, and various AISI 10XX steels. Each material’s performance was evaluated in terms of its tensile strength, weight, and manufacturability. The charts presented the findings of this evaluation, with the chosen material and thickness marked by red circles on the graphs. For the L-bracket, the optimal combination was found to be a ½-inch thick Al 6061 alloy, which provided the necessary strength while maintaining an acceptable weight profile. This decision ensured that the L-bracket would meet the required deflection limits while also adhering to the overall design criteria of cost, manufacturability, and structural performance. The chosen geometry and thickness also contributed to maintaining a sufficient factor of safety, which is a critical parameter in ensuring the long-term integrity of the system.

In addition to the material and thickness considerations, a detailed trade-off analysis was performed to determine the optimal configuration for the adapter. Specifically, the analysis compared the performance of two-hole versus three-hole configurations, as well as the impact of varying adapter lengths. The results of this trade-off analysis were plotted on a chart, which demonstrated how changes in the number of holes and the overall length affected critical parameters such as weight, edge clearance, and mechanical performance. The trade-off chart indicated that the two-hole configuration was the most efficient choice for the adapter, as it provided the required stability and support while minimizing weight. Furthermore, the edge clearance was maximized with the selected length, ensuring that the adapter would properly fit within the spatial constraints of the testbed. The red circle on the chart marked the chosen configuration, which offered the best balance of weight, strength, and clearance, ensuring that the adapter would function effectively while maintaining structural integrity.

The manufacturing process also provided valuable insights into the practical challenges of producing these components. The L-bracket was successfully produced through a combination of external bending and internal machining. For the adapter, the design was optimized for CNC machining, taking into account the available equipment in the on-campus shop. This allowed for a highly accurate and efficient production process, with the final components meeting the necessary tolerances for the experimental setup. The final designs for both the L-bracket and adapter were validated against the requirements, ensuring that they would perform as intended in the testbed assembly.

\section{Conclusion}
The development of the robotic arm testbed has reached a significant milestone, with the successful design, fabrication, and assembly of the key mechanical interfaces that integrate the robotic arm, linear guide-rail system, and optical breadboard table. These interfaces, including the L-bracket, adapter, and 80/20 extrusion table, were carefully engineered to meet the rigid-body assumption and precise tolerances required for simulating ISAM operations. With these critical components now finalized, the next step involves completing the full assembly, ensuring all elements are properly aligned, and securely mounting the robotic arm.

Following this assembly, attention will shift toward refining the system, particularly focusing on tolerances and ensuring that each component operates within the desired specifications. These refinements will address any minor issues that may arise during the integration process, ensuring the system is fully operational and ready for experimental testing. Once the assembly is complete and the system passes final checks, it will be prepared for the next stage of the project, which involves conducting experiments to validate the theoretical models and assess the system’s performance in emulating ISAM operations.

Looking forward, the focus will transition to testing and validation, where the system will be put through a series of trials to evaluate its ability to replicate the dynamics of ISAM tasks. This testing phase will be crucial in determining the effectiveness of the design and providing insights into areas for improvement. Additionally, follow-on work will aim to refine the system further, incorporating feedback from the experimental results and enhancing its functionality for future applications. The testbed will continue to evolve as it undergoes iterative improvements, supporting future research and development in the field of robotic arm-based satellite servicing and ISAM operations.

\bibliographystyle{AAS_publication}   
\bibliography{references}   

\end{document}